\documentclass[10pt,conference,switch]{IEEEtran}
\usepackage{amsmath,amssymb,amsfonts}
\usepackage{algorithm}
\usepackage[noend]{algpseudocode}
\usepackage{graphicx}
\usepackage{textcomp}
\usepackage{xcolor}
\usepackage[frozencache,cachedir=.]{minted}
\usepackage{nccmath}
\usepackage{pifont}
\IEEEoverridecommandlockouts
\IEEEpubid{\makebox[\columnwidth]{Copyright~\copyright2022 IEEE \hfill} \hspace{\columnsep}\makebox[\columnwidth]{ }}

\usepackage{hyperref}
\usepackage{cleveref}
\usepackage{multirow}
\usepackage{multicol}
\usepackage{booktabs}
\usepackage{array}
\usepackage{arydshln}
\usepackage{soul}
\graphicspath{{./img}}
\usepackage{numprint}

\usepackage[style=ieee, citestyle=numeric-comp, backend=bibtex]{biblatex}
\def\BibTeX{{\rm B\kern-.05em{\sc i\kern-.025em b}\kern-.08em
    T\kern-.1667em\lower.7ex\hbox{E}\kern-.125emX}}

\npthousandsep{,}
\npdecimalsign{.}
\newcommand{\changedstrike}[1]{\textcolor{red}{\st{#1}}}
\renewcommand{\changedstrike}[1]{} 

\newcommand{\email}[1]{\href{mailto:#1}{#1}}

\begin{document}

\title{Enriching Source Code with Contextual Data
for Code Completion Models: An Empirical Study}

\author{
\IEEEauthorblockN{Tim van Dam}
\IEEEauthorblockA{\textit{Delft University of Technology} \\
Delft, The Netherlands \\
\email{t.o.vandam@student.tudelft.nl}} 
\and
\IEEEauthorblockN{Maliheh Izadi}
\IEEEauthorblockA{\textit{Delft University of Technology} \\
Delft, The Netherlands \\
\email{m.izadi@tudelft.nl}}
\and
\IEEEauthorblockN{Arie van Deursen}
\IEEEauthorblockA{\textit{Delft University of Technology} \\
Delft, The Netherlands \\
\email{arie.vandeursen@tudelft.nl}}
}

\maketitle

\begin{abstract}
Transformer-based pre-trained models have recently 
achieved great results 
in solving many software engineering tasks
including automatic code completion 
which is a staple in a developer's toolkit. 
While many have striven 
to improve the code-understanding abilities of such models, 
the opposite -- making the code easier to understand -- 
has not been properly investigated.
In this study, we aim to answer 
whether making code easier to understand 
through using contextual data 
improves the performance of 
pre-trained code language models 
for the task of code completion.
We consider \textit{type annotations} and \textit{comments} as
two common forms of additional contextual information 
that often help developers understand code better.
For the experiments,
we study code completion in two granularity levels; 
token and line completion
and take three recent and large-scale language models 
for source code: 
\textit{UniXcoder}, \textit{CodeGPT}, and \textit{InCoder}
with five evaluation metrics.
Finally, we perform the Wilcoxon Signed Rank test 
to gauge significance and measure the effect size.
Contrary to our expectations, 
all models perform better if type annotations are removed 
(albeit the effect sizes are small).
For comments, 
we find that the models perform better 
in the presence of multi-line comments 
(again with small effect sizes).
Based on our observations,
we recommend making proper design choices 
when training, fine-tuning, or simply selecting such models 
given the intended data and application.
Better evaluations and multi-modal techniques 
can also be further investigated 
to improve the practicality and accuracy of auto-completions. 

\end{abstract}

\section{Introduction}
Transformer-based pre-trained models~\cite{Vaswani2017}
originally proposed in the 
Natural Language Processing (NLP) field
have recently been extended to the source code domain~\cite{al2023stacc,izadi2022linkformer,izadi2022catiss}.
Thanks to natural properties of source code~\cite{hindle2016naturalness}
and also the modifications to tailor these models,
they are currently top performers 
in many code-related tasks 
such as automatic code completion 
(hereafter called auto-completion)~\cite{guo2022unixcoder,fried2022incoder,izadi2022codefill,svyatkovskiy2020intellicode}.
Auto-completion techniques
complete source code statements
by suggesting the next token(s)
given the current development context.
They help developers program faster by
correcting typographical errors,
decreasing the typing effort,
and facilitating API exploration~\cite{amann2016study}
making auto-completion one of the most prominent features
in Integrated Development Environments (IDEs).

Auto-completion utilizes two information channels; 
the natural language and the algorithmic channel~\cite{Casalnuovo2020theory}. 
The former explains the context of a program, 
while the latter specifies computer execution.
Comments are a common form of optional information 
that can help developers understand code better,
however, they do not affect how programs are run.
Type annotations are another form of auxiliary information 
to help developers generate and/or understand code better.
They can increase auto-completions' accuracy 
as the type of variables often directly 
dictates how these variables can be interacted with.
However, types are often not present 
in dynamically-typed or optionally-typed languages.
Auto-completion models 
often focus on code tokens 
and lately also on some aspects of program structure 
to provide better completions.  
Most recently, 
comments have also been utilized in such models~\cite{guo2022unixcoder}.
Researchers have also proposed models 
such as LambdaNet~\cite{Wei2020} and TypeBERT~\cite{jesse2021learning}
for the task of type inference.  
However, to what extent 
contextual information embedded 
in source code in the form of  
comments and annotated types 
can impact the performance of 
recent large-scale pre-trained language models 
has not been investigated yet. 

In this work, 
we address this knowledge gap 
by conducting an extensive empirical investigation of the performance of 
recent language models for source code.
We consider the three publicly available models, namely
UniXcoder~\cite{guo2022unixcoder},
CodeGPT~\cite{Lu2021},
and InCoder~\cite{fried2022incoder}.
We perform auto-completion in two granularity levels; 
next-token prediction and line completion.
Moreover, we report the results based on five evaluation metrics
commonly used to evaluate NLP models.

To preserve the underlying semantics of a piece of code,
we add/remove optional auxiliary contextual information,
i.e., type annotations and comments 
and generate multiple variations of the same code. 
To this end, we first collect a dataset containing 
a total of \numprint{704} TypeScript repositories
from the most starred public repositories on GitHub.
We then use the TypeScript compiler to 
create multiple variants of the same TypeScript code.
The first of these variants has all type annotations removed, 
while the second has type annotations added to the code, 
given that the types can be inferred.
TypeScript, being a gradually-typed language, 
does not mandate the presence of type annotations.
The TypeScript compiler is therefore 
equipped with a type inference system 
that can deduce the types of variables without type annotations, 
given that the value of the variable provides enough information.
The three datasets 
are then further processed
by varying the levels of comments.
We consider 
1) keeping comments as-is, 
2) removing all comments, 
3) keeping only single-line comments, 
4) keeping only multi-line comments, 
5) keeping only doc-blocks,
This leads to $15$ datasets containing semantically-similar code, 
however, with different amounts of type annotations and comments.
Then, we use the three models
to perform automatic \textit{token} and \textit{line} completion 
on equivalent versions of TypeScript code with different amounts of type annotations.
This is to establish the effect of 
the presence (or lack thereof) of type annotations.
Additionally, we investigate the effects of removing all comments and limiting the comments 
to only single-line, multi-line, and doc-block comments.

Our results show that 
all three models perform
better on untyped code than on code with type annotations.
To assess the significance of the outperformance,
we conduct the Wilcoxon Signed Rank test.
The $p$-values obtained indicate that the differences
are significant for all models across all evaluation metrics,
meaning the differences are not random.
Note that the effect sizes are small, i.e., 
practical significance may be limited.
Based on the above, 
and considering that all five evaluation metrics 
are mostly affected in the same way, 
further efforts to propose better evaluation metrics 
that assess the value to developers are required.
Additionally, our results indicate that the presence of multi-line
comments significantly contributes to auto-completion performance.
Similar to the previous case, 
although the differences are significant, 
the effect sizes are small in this case as well.
Interestingly, although doc-block comments 
are a type of multi-line comment, 
they do not have a meaningful effect on performance relative to the baseline.

Therefore, the community should take these factors into account 
when selecting the appropriate auto-completion model 
given their purpose and application.
The main contributions of this work are:
\begin{itemize}
    \item An extensive empirical assessment of
    the impact of type information on
    three large language models for both token and line completion in 
    TypeScript code with various amounts of type annotations,
    \item A comprehensive empirical assessment of 
    the effect of natural language text information 
    in three formats (single-line, multi-line, and doc-block comments)
    on the performance of these code language models 
    for both token and line completion,
    \item Our source code, dataset, 
    and select fine-tuned models are publicly available.~\footnote{\url{https://github.com/AISE-TUDelft/ContextualDataCodeCompletion}; \url{https://huggingface.co/AISE-TUDelft/CodeGPT-TS-Multi-Untyped}; \url{https://huggingface.co/AISE-TUDelft/UniXcoder-TS-Multi-Untyped}}
\end{itemize}

\section{Motivating Example}\label{sec:motiv}
Figure~\ref{fig:motivating_example} 
shows an example code snippet from the Angular repository
with and without type annotations and comments.\footnote{\href{https://github.com/angular/angular/blob/8d485491a9630e1b24e1781b98335f011764699a/packages/router/src/utils/tree.ts\#L78}{https://github.com/angular/.../\#L7}} 
Lack of appropriate type information or documentation 
can make it harder for developers to use this function properly. 
For instance, one might provide a string as \texttt{value} 
but a tree of numbers as \texttt{node}.
In statically-typed languages,
this would lead to a compile-time error, 
however, in dynamically-typed languages, 
this would run perfectly fine, 
which can lead to bugs entering the code base.
It is therefore important for auto-completion models 
to work well in situations 
where type annotations are lacking 
to prevent users from introducing bugs 
to their code through auto-completion.
In theory, additional type information should boost auto-completion models, 
as it provides them with a more comprehensive description of the source code.
The same can be said about comments, 
which are typically used to describe complete functions with doc-blocks, 
or used to annotate smaller parts of code with single-line comments.
The difference between the two is that 
type annotations are placed in a structured manner, 
whereas comments are not guaranteed to follow a specific structure. 
Note that doc-blocks do follow a form of structure, 
but do not have an order and can contain a wide range of information.
Investigating the impact of types and comments and their relationship
on the performance of pre-trained Language Models (LMs) for auto-completion 
in both dynamically- and gradually-typed languages 
with varying amounts of type annotations
gives us an understanding into 
what elements of source code can be used 
to improve the performance of auto-completion approaches.

\begin{figure}[tb]
    \begin{minted}[frame=single,linenos,fontsize=\footnotesize,breaklines,breaksymbol={},numbersep=1mm]{typescript}
// Return the path to the node with the given value using DFS
function findPath<T>(value: T, node: TreeNode<T>): TreeNode<T>[] {
    if (value === node.value)
        return [node];
    for (const child of node.children) {
        const path: TreeNode<T>[] = findPath(value, child);
        if (path.length) {
            path.unshift(node);
            return path;
        }
    }
    return [];
}
    \end{minted}
    \begin{minted}[frame=single,linenos,fontsize=\footnotesize,breaklines,breaksymbol={},numbersep=1mm]{javascript}
function findPath(value, node) {
    if (value === node.value)
        return [node];
    for (const child of node.children) {
        const path = findPath(value, child);
        if (path.length) {
            path.unshift(node);
            return path;
        }
    }
    return [];
}
    \end{minted}
    \caption{Sample code snippet with and without type annotations and comments}
    \label{fig:motivating_example}
\end{figure}

\section{Background and Related Work}\label{sec:related}
In the following, 
we first provide background
on pre-trained models, 
then we review the existing work on completing code 
using these approaches. 

\subsubsection{Transformers and Pre-trained Models}
Transformer-based~\cite{Vaswani2017} models 
have recently shown great promise in the area of NLP.
BERT~\cite{Devlin2018}, a bidirectional Transformer model, 
showed the value of considering both the left and right context 
for training LMs.
\citeauthor*{Liu2019} improved further upon BERT 
with the RoBERTa~\cite{Liu2019} model, 
and aimed to show that the performance of BERT 
can be further be improved 
through optimizing different design choices.
Raffel et al.~\cite{Raffel2019} 
proposed a Text-to-Text Transformer (T5) 
which treated various NLP goals as a \textit{text-to-text} 
(i.e., seq2seq) task.
While BERT-based models use \textit{Masked} Language Modeling (MLM), 
the Generative Pretraining Transformer (GPT) architecture 
utilizes \textit{Causal} Language Modeling (CLM)
and is suitable for generation tasks~\cite{Radford2019}.
Nowadays, Transformers are being tailored to
source code to solve software engineering tasks.
For instance, \citeauthor*{Feng2020}'s \textit{CodeBERT}~\cite{Feng2020}, 
builds on top of the RoBERTa model introduced above.
\textit{CodeT5} is the corresponding T5 model 
for the source code fine-tuned on multiple code-related tasks 
such as code summarisation, translation, generation, and more~\cite{Wang2021,al2023extending}.
\textit{Codex} is an LM for code that is based on the GPT-3 architecture~\cite{chen2021evaluating, brown2020language}.
The authors show that Codex is capable of implementing full-function implementations from textual prompts and function signatures alone.
Although most research on auto-completion
is focused on single-token prediction,
several studies aimed to 
complete entire statements or blocks of code~\cite{svyatkovskiy2020intellicode,nguyen2019combining,yang2017language}.
For instance, 
\textit{AUTOSC} combines program analysis and software naturalness
and fills in a partially completed statement
with frequent and valid recommendations~\cite{nguyen2019combining}.
\textit{GPT-C} is a multi-lingual model based on GPT-2,
for completing lines~\cite{svyatkovskiy2020intellicode}.
\textit{CodeFill} is Multi-Task Learning-based approach also based on GPT-2 
for completing lines for dynamically-typed languages~\cite{izadi2022codefill}.
The authors showed using the extra information 
from the structure representation 
is beneficial for the model.

\subsubsection{Studies on Type Annotations and Comments}
Several previous works have demonstrated the ability 
to infer the types of variables and functions in dynamically-typed languages 
depending on their context~\cite{Malik2019, Hellendoorn2018, Wei2020, jesse2021learning}.
However, whether type information can be used to improve code understanding has not been established.
\textit{DeepTyper}~\cite{Hellendoorn2018} 
shows how deep learning can be applied 
to infer types to ease the transition from untyped code to gradually-typed code.
Similar to DeepTyper~\cite{Hellendoorn2018},
\citeauthor*{Malik2019}'s NL2Type~\cite{Malik2019} 
shows that natural language information in comments, functions, and parameter names 
can be exploited to predict types in dynamically-typed languages.
\citeauthor*{Wei2020}'s \textit{LambdaNet}~\cite{Wei2020} 
shows that deep learning can be applied to provide untyped code 
with type annotations in gradually-typed languages like TypeScript and Python.
LambdaNet is able to predict user and third-party types, 
while DeepTyper is only able to predict types from a fixed vocabulary.
Similar to LambdaNet~\cite{Wei2020}, 
\textit{TypeBERT}~\cite{jesse2021learning} 
is able to infer user and third-party types
but it takes a much simpler approach 
by applying BERT-style pre-training, 
after which it is fine-tuned on a large set of TypeScript data.
\citeauthor*{mastropaolo2021empirical} 
compare T5~\cite{Raffel2019} 
to n-gram models on a comment completion task, 
showing T5~\cite{Raffel2019}, 
 leverages code context 
to complete partial comments~\cite{mastropaolo2021empirical}.
As previously mentioned, \textit{Codex}~\cite{chen2021evaluating} has shown proficiency in generating function implementations from natural language descriptions and a function signature alone. This shows that natural language can be of value to LMs for code.
Additionally, these models show that it is possible for LMs to infer type information from source code.
However, the opposite, whether type information can be leveraged to facilitate code understanding, has not been established.
UniXcoder is a Transformer-based model proposed by \citeauthor*{guo2022unixcoder}~\cite{guo2022unixcoder}.
Guo et al. show that UniXcoder performs slightly better 
on the auto-completion task when considering comments. 
However, the impact of type annotations and different types of comments,
i.e., single-line and multi-line comments 
are not considered individually.
Similarly, \citeauthor*{fried2022incoder}~\cite{fried2022incoder}
show good single-token prediction performance, but do not investigate
the performance of line completion, 
nor do they consider the influence of comments or type annotations.
Note that we adapt both these models to perform line completion 
(more details in the approach section).

\subsubsection{Empirical Studies on Auto-Completion Models}
\citeauthor*{Ciniselli2021}~\cite{Ciniselli2021, Ciniselli2021_2} 
analyzed the performance of two language models for text namely, 
T5~\cite{Raffel2019} and RoBERTa~\cite{Liu2019},
for completing code in three granularity levels; 
single-token, line, and block.
The authors included two datasets, 
containing Java methods and Android app methods 
from open-source GitHub repositories.
They showed that T5 performs better,
however, the success of these models 
when tasked to predict longer sequences is limited.
As only Java was used for evaluation,
the results are not generalizable 
to dynamically- or gradually-typed languages.
Similar to our study,
the authors aim to assess Transformer models' performance,
however, the angles of their study differ from this study.
While they focus on investigating 
the prediction granularity levels for T5 and RoBERTa,
we focus on the impact of comments and annotated types on
the performance of state-of-the-art large-scale LMs 
specifically fine-tuned on \textit{source code}.
\citeauthor*{chirkova2021empirical}~\cite{chirkova2021empirical} 
analyze the performance of Transformers on several code-related tasks including auto-completion.
This work analyzes how well Transformer models are able to perform tasks using solely syntactic information.
They show that the auto-completion task uses all AST components and that omitting types in ASTs
has a negative impact on this task.

\section{Study Design} \label{sec:methodology}
\begin{figure*}[tb]
\centering
\includegraphics[width=1.5\columnwidth]{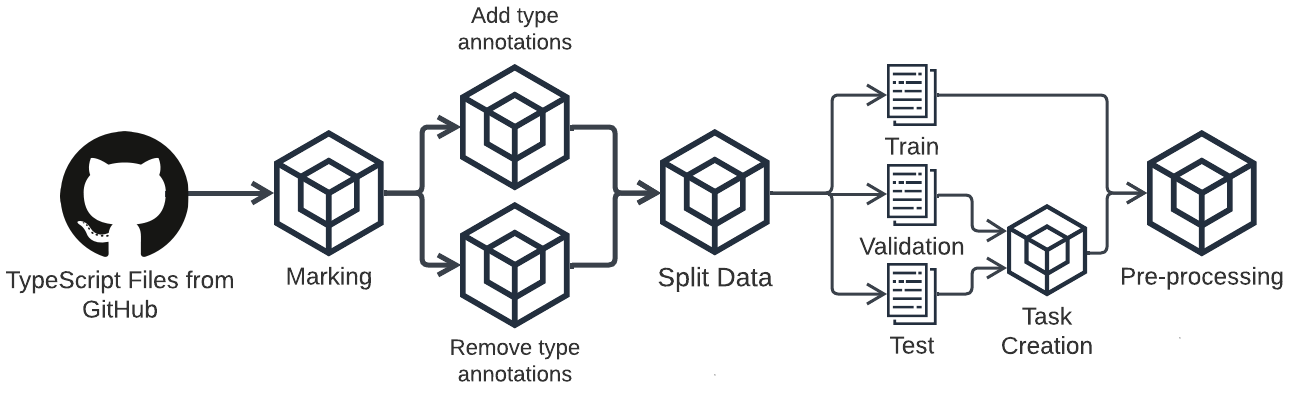}
\includegraphics[width=1.5\columnwidth]{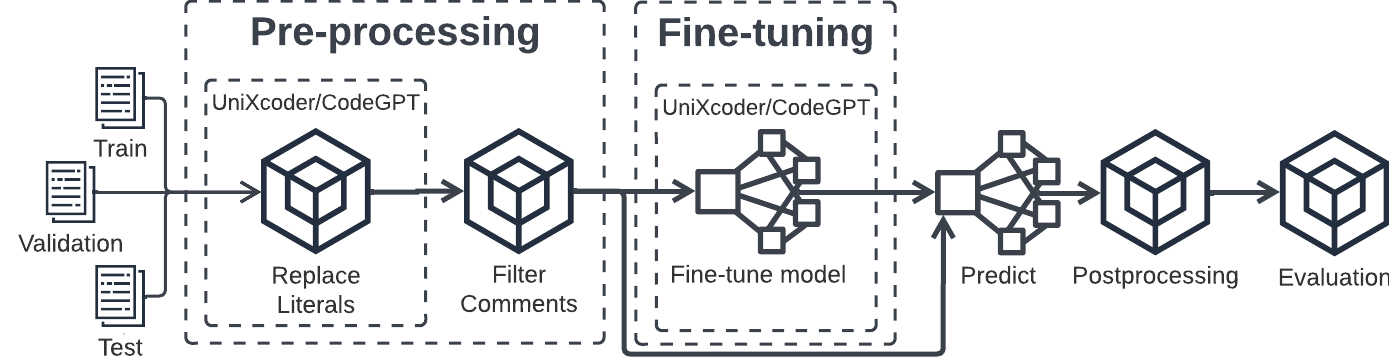}
\caption{Overview of the study pipeline}
\label{fig:full_pipeline}
\end{figure*}

We select three of the most recently released LMs for source code 
that are publicly available, namely UniXcoder, CodeGPT, and InCoder.
Choosing three models leads to
more generalizable results across the research questions.
Moreover, these models support two different language modeling objectives; 
masked and causal. 

\textit{UniXcoder} is a pre-trained model 
that leverages multiple modalities 
to facilitate several code understanding and generation tasks~\cite{guo2022unixcoder}.
In addition to source code, 
comments and flattened Abstract Syntax Trees (ASTs) 
were used during pre-training 
to improve understanding.
UniXcoder was pre-trained using 
MLM~\cite{Devlin2018, Baevski2019}, 
Unidirectional Language Modeling (ULM)~\cite{Radford2018},
and denoising objectives~\cite{Raffel2019}.
Initially, UniXcoder was trained on unimodal natural language data from the C4~\cite{Raffel2019} dataset. 
Afterward, it was trained on bimodal data 
in the form of text-code pairs from the CodeSearchNet dataset~\cite{Husain2019}.
These pairs consist of function definitions and corresponding leading comments.

\textit{InCoder} is a Transformer-based decoder-only model
which is able to infill code based on both left and right contexts~\cite{fried2022incoder} 
using causal modeling~\cite{Aghajanyan2022}.
It was trained 
on a large dataset consisting primarily of Python and JavaScript code
from GitHub, GitLab, and StackOverflow.
The training data from StackOverflow contains natural text
from questions, answers, and comments
in addition to code.

\textit{CodeGPT} is a Transformer-based model based on the GPT-2 model.
\citeauthor*{Lu2021} utilize the GPT-2 architecture 
to train several LMs for code-related tasks.
The authors train four models, two for Python and two for Java: 
for each programming language,
one model solely used the GPT-2 architecture, 
and one used the pre-trained text-based GPT-2 as a starting checkpoint.
The latter methodology was used for auto-completion by \citeauthor*{Lu2021}
and is consequently what is referred to as CodeGPT in this work.

Our study consists of seven main phases, namely
\textit{marking},
\textit{type inference},
\textit{data splitting},
\textit{pre-processing},
\textit{hyper-parameter tuning},
\textit{fine-tuning},
and \textit{post-processing}.
During marking, we add comments to the TypeScript code 
in our dataset to determine where we will perform auto-completion.
Afterward, we create equivalent marked datasets with 
1) all type annotations removed, 
2) all implicit types added, and
3) where type annotations are left as-is.
For all three datasets, we divide the files into train, test, and validation sets 
according to an $80/10/10$ percent split.
There are 5 pre-processing modes that all handle comments differently: 
1) keep all comments, 
2) remove all comments, 
3) keep only multi-line comments, 
4) keep only single-line comments, and 
5) keep only doc-blocks.
This results in $15$ unique datasets.
Next, we perform hyper-parameter tuning, 
after which we use these datasets to fine-tune $15$ models for UniXcoder and CodeGPT.
Afterward, we use the fine-tuned models 
to generate predictions for the test set. 
Finally, we post-process the predictions.

Note that we do not fine-tune InCoder: it does not have publicly available fine-tune code and has already been trained on data containing TypeScript code.
This also has an effect on the pre-processing step for InCoder, as discussed later.

Figure~\ref{fig:full_pipeline} depicts the overall pipeline of our study.
Note that different paths are taken depending on the auto-completion model in use.
Next, we describe the steps in more detail.

\paragraph{Marking}
When testing the fine-tuned models,
marking indicates where auto-completion 
should be performed.
Throughout a piece of code, 
we add placeholder comments in the form of \texttt{/*<marker:i>*/}, 
where $i$ is a number that we use to identify markers within a file.
These comments are later used to create the 
auto-completion tasks for the test and validation sets.
We perform this task prior to other phases 
to ensure that each dataset has identical completion tasks:
comments will not move as we transform code, 
hence we can use them to track a position in code even when transformations 
occur during type inference and pre-processing.
For marking, we first randomly select $40\%$ of non-empty lines.
Then, for each selected line, 
we place a marker in front of an ECMAScript token 
that is not a \textit{newline}, \textit{whitespace}, \textit{comment}, or \textit{type annotation}.
The code to the left of this marker will serve as input, 
and the code to the right of this marker 
up to the end of the line 
will serve as ground truth.

\paragraph{Type Inference}
In order to ascertain the influence of the number of types on performance, we 
transform our TypeScript dataset in order to create two additional datasets:
one of which makes all implicit types explicit, and one of which is without any type annotations.
In the former case, we for instance transform \texttt{const x = 5 * 10;} to \texttt{const x: number = 5 * 10;}.
The opposite transformation would happen in the latter case.
This results in three datasets, with three different amounts of type annotations.
This step is more elaborately discussed in \autoref{sec:experminets}.

\paragraph{Data splitting}
We split all TypeScript files into the train, test, and validation sets.
All three datasets contain the same files,
so we maintain one distribution for all datasets.
This ensures that all models produce comparable results.
The files are split according to an $80/10/10$ percent split.

\paragraph{Pre-Processing}
We pre-process the data 
to prepare for fine-tuning and evaluation.
First, we normalize spacing and linebreaks by replacing consecutive spaces and linebreaks with a single such character.
Then, the following steps are applied
to prepare the data for fine-tuning UniXcoder and CodeGPT:
As a standard technique 
to reduce the vocabulary size in the literature~\cite{izadi2022codefill,svyatkovskiy2020intellicode,izadi2021topic,izadi2022predicting},
we normalize the number and string literals 
by replacing them with special tokens, 
$\left<\text{NUM\_LIT}\right>$, 
and $\left<\text{STR\_LIT}\right>$, respectively.
Then, we replace line breaks 
with the special $\left<\text{EOL}\right>$ token.
These pre-processing steps are not applied to the data that is fed to InCoder,
as we do not fine-tune this model.
Consequently, the InCoder expects raw code, 
without the special tokens that the pre-processing phase adds.

%
We then apply different variants of pre-processing regarding comments;
1) keep all comments as-is,
2) remove all comments,
3) keep only single-line comments,
4) keep only multi-line comments, and finally
5) keep only doc-blocks.
This process results in $15$ different collections of TypeScript files 
(three type variants, and five comment variants).

The pre-processed data is then stored in files to be used by the three models.
To prevent large files from representing a large part of the validation and test sets,
the maximum amount of code completion tasks per file is set at $15$.

\paragraph{Hyper-parameter Tuning}
We perform hyper-parameter tuning on
$25\%$ of the training and validation set of the dataset 
with the original types (TS704-OT) and unmodified comments 
to perform hyper-parameter tuning.
We choose to use 25\% of one dataset to limit the computational burden of our experiments.
Note that fine-tuning \textit{does} use 100\% of all datasets.
Table~\ref{table:hyperparams} reports the tuned hyper-parameters, 
the values that were tested, and the combination of hyper-parameters 
which led to the highest accuracy.
\begin{table}[tb]
\caption{Tuned Hyper-parameters}
\label{table:hyperparams}
\centering
\begin{tabular}{llll}
\toprule
Model & Hyper-param & Values & Best \\
\midrule
\multirow{2}{*}{UniXcoder}
& Learning Rate & $\left\{ 1\text{e-}5, 7.33\text{e-}5, 1.37\text{e-}4, 2\text{e-}4 \right\}$ & 7.33\text{e-}5 \\
& Batch Size & $\left\{ 2, 4, 8 \right\}$ & 4 \\

\cdashline{2-4}\noalign{\vskip \belowrulesep}

\multirow{2}{*}{CodeGPT}
& Learning Rate & $\left\{ 1\text{e-}5, 7.33\text{e-}5, 1.37\text{e-}4, 2\text{e-}4 \right\}$ & 1.37\text{e-}4 \\
& Batch Size & $\left\{ 2, 4, 8 \right\}$ & 2 \\
\bottomrule
\end{tabular}
\end{table}

\paragraph{Fine-tuning}
Next, we fine-tune UniXcoder and CodeGPT on all our datasets.
That is, on each type-annotation variant and each comment-variant.
This results in 15 different datasets, thus 15 different fine-tuned models for UniXcoder and CodeGPT.
We use the hyper-parameters found during hyper-parameter tuning.
Note that we do not fine-tune InCoder, 
as it does not have publicly available fine-tune code.
Furthermore, it was already trained on TypeScript code, 
hence it is expected to be able to provide reasonable predictions on the test set without fine-tuning.
We use a standard language modeling objective, 
predicting the next token given a context,
and maximize the following likelihood.
In Equation~\ref{eq:log_statement},
$m$ is the length of the predicted sequence of code token values
and $\theta$ is the set of parameters
that is learned through
stochastic gradient descent optimization 
to model $P$~\cite{robbins1951stochastic}.

\begin{equation}
    L(V) = \sum_{i}{\log{P(v_i | c_0,...,c_T,v_{i-m},...,v_{i-1};\theta)}}.
    \label{eq:log_statement}
\end{equation}
UniXcoder and CodeGPT treat the $\left<\text{EOL}\right>$ tokens 
as the end of a sequence token, 
resulting in the fine-tuned models predicting up to the end of each line.
To emulate this behavior in InCoder, 
we add a stopping criterion to the model that detects 
whenever a new line is among the generated tokens.
If a new line is detected, 
we stop the model from generating more tokens.

\paragraph{Post-processing}
After fine-tuning, the $30$ fine-tuned models plus InCoder 
are used to generate predictions for the $15$ test sets.
These predictions are subsequently post-processed 
which consists of normalizing 
the spacing of code tokens (including line breaks), 
removing all comments, 
and replacing tokenized versions of literals 
with default literals.
That is, $\left<\text{STR\_LIT}\right>$ 
becomes \texttt{""} (an empty string), 
and $\left<\text{NUM\_LIT}\right>$ becomes \texttt{0}.
Since the data fed to InCoder 
does not contain tokenized versions of literals, 
we replace the raw string and number literals 
in the InCoder's predictions by these constants.
To have a consistent and fair evaluation, 
this process is applied 
to both the prediction and the ground truth.

\section{Experimental Setup}\label{sec:experminets}
We first present our Research Questions (RQ)
and describe the datasets 
used for the experiments.
Next, we introduce the evaluation metrics 
used to assess the models' performance in detail.
Finally, we review the implementation details.

\subsection{Research Questions}
We aim to assess the impact of additional contextual information,
e.g., explicit type annotations and textual explanations 
in the form of comments 
on the performance of three state-of-the-art large-scale 
pre-trained models for source code.
For both of our RQs, 
we consider three code LMs, i.e., 
UniXcoder, CodeGPT, and InCoder, 
and two auto-completion tasks, namely
token and line completion.
Accordingly, we design our experiments to 
answer the following RQs.
\begin{itemize}
    \item \textbf{RQ1: How is the performance of these models 
    influenced by the ratio of available type annotations in code?}
    That is, whether these models perform significantly differently 
    depending on the degree to which a piece of code is type-annotated.
    To provide a fair comparison, 
    we use a TypeScript dataset,
    and create its equivalent code without type annotations 
    using the TypeScript compiler. 
    Moreover, to assess whether adding additional type annotations 
    affect the performance of models,
    we add more annotations to the TypeScript dataset 
    to make it more explicitly typed
    and run the same experiments on this new dataset.
    \item \textbf{RQ2: What is the impact of enriching 
    source code context 
    with textual information, i.e., comments?}
    This question explores 
    how the performance of the three models
    is affected by the presence of different types of comments
    including single-line, multi-line, and doc-block comments.
    We compare these results against 
    the results obtained from 
    the respective datasets with no comments.
\end{itemize}

\subsection{Datasets}
\begin{figure}[tb]
    \begin{minted}[frame=single,linenos,fontsize=\footnotesize,breaklines,breaksymbol={},numbersep=1mm]{typescript}
function solveQuadratic(a: number, b: number, c: number) {
    const d = b ** 2 - 4 * a * c;
    const denom = 2 * a;
    const sol1 = (-b + Math.sqrt(d)) / denom;
    const sol2 = (-b - Math.sqrt(d)) / denom;
    return [sol1, sol2];
}
    \end{minted}
    \begin{minted}[frame=single,linenos,fontsize=\footnotesize,breaklines,breaksymbol={},numbersep=1mm]{typescript}
function solveQuadratic(a: number, b: number, c: number): number[] {
    const d: number = b ** 2 - 4 * a * c;
    const denom: number = 2 * a;
    const sol1: number = (-b + Math.sqrt(d)) / denom;
    const sol2: number = (-b - Math.sqrt(d)) / denom;
    return [sol1, sol2];
}
    \end{minted}
    \caption{A sample code snippet before/after adding additional type annotations}
    \label{fig:code_adding_annotations}
\end{figure}

To perform the experiments, 
we use publicly available GitHub repositories 
that predominantly consist of TypeScript code.
\textbf{TS704-OT}, the first dataset, 
consists of a subset of the top-1000 starred repositories on GitHub.
This dataset was retrieved by querying the 
GitHub Search API.\footnote{\href{https://docs.github.com/en/rest/search?apiVersion=2022-11-28\#search-repositories}{https://docs.github.com/en/rest/search?apiVersion=2022-1..repositories}}
The GitHub API returned a total of \numprint{851} unique repositories.
To prevent bias, 
we deduplicate our dataset against the repositories used for (pre-)training UniXcoder, CodeGPT, and InCoder.
To answer the first RQ,
we created an additional dataset
based on the TS704-OT dataset
where we remove all type annotations (\textbf{TS704-NT}).
To do so, we use the TypeScript Compiler API to traverse the Abstract Syntax Tree of every TypeScript file, removing any type annotations that are encountered.
Additionally, we created a dataset
where we make all implicit types explicit 
using the TypeScript compiler (\textbf{TS704-AT}).
As TypeScript is a gradually-typed language, 
types are not required but can oftentimes be inferred based on the types of other variables or constants.
We use the TypeScript compiler to add type annotations when they can be inferred by the compiler, attempting to amplify a potential effect caused by the presence of type annotations.
This process is displayed with a sample code snippet in Figure~\ref{fig:code_adding_annotations}.
TypeScript code may depend on type annotations that are defined in third-party dependencies.
Hence, we first install all third-party dependencies using \textit{npm} (Node Package Manager) to be able to infer these types.
More specifically, we run the \texttt{npm install --ignore-scripts} command in each directory containing a \texttt{package.json} file (\texttt{package.json} files contain dependency information).
We opt to ignore post-install scripts, as \texttt{npm} packages made for TypeScript ship with type declarations as-is, meaning that no additional scripts are required to retrieve all third-party types.
This also significantly speeds up the installation process.
Not all dependencies were available, hence we removed all projects with unavailable dependencies from all 
three datasets.
This decreased the number of repositories in our datasets by 147.
After having installed the dependencies, we locate all directories containing \texttt{tsconfig.json} files.
These directories are 
at the root of TypeScript projects, and contain configuration options for the TypeScript compiler.
We use the TypeScript compiler to load in these TypeScript projects and traverse the AST of each TypeScript file (files with the \texttt{.ts} extension), 
adding type annotations where possible using the TypeScript Compiler Type Checker.
The resulting dataset 
The datasets are nearly identical: the sole difference is the number of type annotations.

Table~\ref{table:datasets} 
shows the size of our datasets in terms of 
the number of repositories, files, 
and lines of code (LOC).
Additionally, it displays the \textit{type explicitness} 
of the code in these datasets.
Type explicitness refers to 
the number of type annotations present in the code 
relative to the maximum amount of type annotations
that can be present in the code.
\begin{table}[tb]
\caption{Datasets used for fine-tuning and evaluation}
\label{table:datasets}
\centering
\begin{tabular}{lrrr}
\toprule
& TS704-OT & TS704-NT & TS704-AT\\
\midrule
\#Repositories & \numprint{704} & \numprint{704} & \numprint{704}\\
\#Files & \numprint{174500} & \numprint{174500} & \numprint{174500}\\
\#LOC & \numprint{26115719} & \numprint{25548595} & \numprint{26115719}\\
Type Explicitness & \numprint{30.95}\% & \numprint{0.00}\% & \numprint{95.62}\%\\
\bottomrule
\end{tabular}
\end{table}

\subsection{Evaluation Metrics}
We compare the predictions made by the models, fine-tuned on typed and non-typed code,
against the ground truths using several evaluation metrics.
In this study, we include a wide range of 
standard metrics commonly used 
for evaluating line completion solutions 
to provide a more comprehensive assessment~\cite{izadi2022evaluation,izadi2022codefill,svyatkovskiy2020intellicode}.
As our focus is on assessing the performance 
of these models 
in various settings,  
we review these metrics in detail.

\textbf{EM} (Exact Match) compares ground truths with predictions and returns a boolean value.
The EM score over an entire dataset is expressed as a percentage.
Higher values are better.

\textbf{ES} (Edit Similarity) or Levenshtein Similarity 
compares the ground truth 
to the prediction on a character-by-character basis.
Wrong characters (substitutions), 
too many characters (insertions) 
and too few characters (deletions) 
increase the Levenshtein distance by $1$.
ES is a number 
in the range $\left[0, 1\right]$ 
that is computed 
by dividing the Levenshtein distance 
by the length of either the prediction or the ground truth, 
depending on which is the longest.

\textbf{BLEU-4} is a variant of BLEU (Bilingual Evaluation Understudy) 
that deals specifically with n-grams 
with $\text{n} \in \left[1, 4\right]$.
BLEU compares the ground truth to the prediction 
by computing the ratio of n-grams 
that occur in both the prediction and the ground truth 
to the total amount of n-grams in the ground truth~\cite{Papineni2002bleu}.
We apply BLEU-4 
by treating each code token, 
as per the ECMAScript lexical grammar specification~\cite{ESTokenSpec}, 
as a unigram, 
and give each value of $n$ an equal weight.
A smoothing technique~\cite{Lin2004orange} 
is used to prevent division by zero 
when the prediction has fewer than four tokens.

\textbf{ROUGE-L} is a variant of the ROUGE 
(Recall-Oriented Understudy for Gisting Evaluation) metric which
uses the Longest Common Subsequence algorithm 
to find the largest n-gram that occurs 
in both the prediction and the ground truth.
ROUGE-L computes precision and recall 
and uses them to compute an F1-score~\cite{Lin2004text}.

\textbf{METEOR} (Metric for Evaluation of Translation with Explicit ORdering), 
compares the ground truth and the prediction 
by mapping their respective unigrams, 
and computes a score over this mapping 
based on its Precision and Recall.
Additionally, a penalty is applied based 
on how well the true unigram order is followed by the mapping.
METEOR has been shown to be better 
at capturing human judgment 
over a complete dataset than BLEU~\cite{Banerjee2005}.
Opposed to BLEU, METEOR puts 
more weight on recall, 
which has been shown to align closer 
to human judgment than precision~\cite{Lavie2004}.
We apply METEOR with parameters 
$\alpha = 0.9$, $\beta = 3.0$, and $\gamma = 0.5$.

\subsection{Implementations and Configuration}
We use \texttt{ts-morph}
to interface with the TypeScript compiler API 
when removing or adding type annotations.\footnote{\url{https://www.npmjs.com/package/ts-morph/v/15.1.0}}
We use \texttt{js-tokens} 
for tokenizing TypeScript code 
during the line completion task creation, pre-processing, and evaluation phases
to add and remove specific types of comments.\footnote{\url{https://www.npmjs.com/package/js-tokens/v/8.0.0}}

We fine-tune UniXcoder using the fine-tune source code published by its authors.\footnote{\href{https://github.com/microsoft/CodeBERT/tree/master/UniXcoder/downstream-tasks/code-completion}{https://github.com/microsoft/CodeBERT/tree/master/UniXcoder}}
We also fine-tune CodeGPT using the published code.\footnote{\href{https://github.com/microsoft/CodeXGLUE/tree/main/Code-Code/CodeCompletion-line/}{https://github.com/microsoft/CodeXGLUE/.../CodeCompletion-line}}
Slight tweaks were made to both scripts 
to make them compatible with our dataset files.

For UniXcoder, we use an initial learning rate of $7.33\text{e-}5$ 
and a batch size of $4$, 
as per the results of hyperparameter tuning (Table~\ref{table:hyperparams}).
We set the maximum input sequence length to $936$, 
the maximum output sequence length to $64$, 
the beam size to $5$, 
The remaining parameters are set 
to the default values as per the fine-tuning code 
published by the authors of UniXcoder.
We then train UniXcoder for $10$ epochs.
For CodeGPT, we use an initial learning rate of $1.37\text{e-}4$ 
and a batch size of $2$ (Table~\ref{table:hyperparams}).
The remaining parameters are set to the default values.
We then train CodeGPT for $10$ epochs.

We do not fine-tune InCoder, as it does not have publicly available fine-tune code.
We use the InCoder model with 1 billion parameters, which is available on HuggingFace.\footnote{\url{https://huggingface.co/facebook/incoder-1B}}
While InCoder is capable of providing completions given left \textit{and right} context, we choose to only provide a left context as input, because UniXcoder and CodeGPT only support left contexts.
For inference, we apply InCoder with a temperature of $0.2$, and a $p = 0.95$ for top-p nucleus sampling.

We conducted the experiments with models
on a cluster equipped with NVIDIA Tesla V100S GPUs, 
an AMD EPYC 7402 24C @ 2.80GHz CPUs.
We ran hyper-parameter tuning, fine-tuning, and inference on this cluster 
using one GPU per process, four CPUs per process, and $48$ GB RAM per process.

Running hyper-parameter tuning on UniXcoder and CodeGPT took around $34$ hours per model.
Fine-tuning these models took roughly $5$ days per model.
Inference on the test set took roughly $6$ hours for every model, including InCoder.

\section{Results and Discussion}\label{sec:results}

\subsection{RQ1: Impact of Type Annotations}
To gauge the impact of type annotations on auto-completion performance,
we run the three models against the test sets of datasets with different type explicitness ratios.
We then report the results for each task, 
namely line and token completion.
Next, to determine which performance difference is significant
we perform the Wilcoxon Signed Rank test.
In the following tables,
we abbreviate BLEU-4, ROUGE-L, and METEOR metrics to
B4, RL, and MR, respectively.

\subsubsection{Line Completion}

\begin{table}[tb]
\caption{RQ1: Impact of Type Annotations (Line Completion, data without comments)}
\label{table:rq1_metrics}
\centering
\begin{tabular}{llrrrrr}
\toprule
Model & Types & EM & ES & B4 & RL & MR\\
\midrule
\multirow{3}{*}{UniXcoder}
& NT & \textbf{\numprint{65.32}} & \textbf{\numprint{79.62}} & \textbf{\numprint{61.63}} & \textbf{\numprint{81.32}} & \textbf{\numprint{65.91}}\\
& OT & \numprint{58.85} & \numprint{73.68} & \numprint{58.25} & \numprint{75.83} & \numprint{63.05}\\
& AT & \numprint{59.93} & \numprint{74.41} & \numprint{58.82} & \numprint{76.39} & \numprint{63.56}\\
\midrule
\multirow{3}{*}{CodeGPT}
& NT & \textbf{\numprint{63.39}} & \textbf{\numprint{80.38}} & \textbf{\numprint{62.08}} & \textbf{\numprint{82.51}} & \numprint{67.03}\\
& OT & \numprint{60.29} & \numprint{78.00} & \numprint{61.30} & \numprint{80.63} & \numprint{66.80}\\
& AT & \numprint{61.32} & \numprint{78.74} & \numprint{61.86} & \numprint{81.13} & \textbf{\numprint{67.25}}\\
\midrule
\multirow{3}{*}{InCoder}
& NT & \textbf{\numprint{33.26}} & \textbf{\numprint{56.56}} & \textbf{\numprint{43.51}} & \textbf{\numprint{59.71}} & \textbf{\numprint{51.09}}\\
& OT & \numprint{32.00} & \numprint{54.99} & \numprint{42.70} & \numprint{58.15} & \numprint{50.04}\\
& AT & \numprint{32.01} & \numprint{54.76} & \numprint{42.36} & \numprint{57.82} & \numprint{49.77}\\
\bottomrule
\end{tabular}
\end{table}

Table~\ref{table:rq1_metrics} 
presents how each model performed 
on comment-less code with and without type annotations.
The displayed data is retrieved by running the models against the test set, computing every metric for each prediction, and averaging the metrics.
We then use the Wilcoxon Signed Rank test to determine whether different type explicitness rates significantly impact auto-completion performance.
Specifically, we compare TS704-NT with TS704-OT, and TS704-OT with TS704-AT.
The test sets contain a total of \numprint{153147} paired code completion tasks.
We obtain paired samples from these tasks by pairing the metric values
computed for predictions done by the two models under test.
This is done for every metric.
The family-wise significance level of the tests is $\alpha=0.05$.
Additionally, we compute Cliff's delta to measure the effect size.
Table~\ref{table:results_rq1_p_values}
reports partial results of the statistical tests.
The B4, RL, and MR metrics are omitted for brevity,
as these were significant in all cases.

\textbf{Removing type annotations} leads to the best performance for all three models. In all cases, this outperformance is statistically significant but quite small -- especially for InCoder.
This performance difference may be explained by the fact that UniXcoder and InCoder are pre-trained on corpora that include large amounts of JavaScript, which is nearly identical to TypeScript without type annotations.
However, it should be noted that even CodeGPT, which does not have any source code in its pre-training data, also shows better performance on code without type annotations.
This opposes the rationale that auto-completion performance can be enhanced by adding type annotations.
A possible explanation could be that source code tokens are more valuable to the models than type annotations, and removing type annotations leaves more input space for source code tokens.

\textbf{Adding type annotations} to partially-typed code improves auto-completion, but not as much as removing type annotations altogether.
For UniXcoder and CodeGPT, adding type annotations led to a statistically significant performance improvement for all metrics.
For InCoder, a similar improvement was observed for all metrics except for Exact Match.
In all cases the effect size is much smaller than the effect size caused by removing all type annotations, suggesting limited practical utility.
While this can not be determined from our experiments alone, these findings may suggest that strongly typed languages are easier to interpret than gradually-typed languages such as TypeScript.
However, the fact that removing type annotations led to a bigger performance increase once more suggests that source code may be a more important part of the input than type annotations.

\begin{table}[tb]
\caption{RQ1: Impact of Type Annotations (Line Completion, data without comments, 
Wilcoxon Signed Rank $p$-values, Cliff's Delta)}
\label{table:results_rq1_p_values}
\centering
\begin{tabular}{llllrr}
\toprule
Model & Types 1 & Types 2 & Metric & $p$ & $\delta$ \\
\midrule
\multirow{4.4}{*}{UniXcoder}
& NT & OT & EM & \numprint{0.000} & \numprint{0.065} \\
& NT & OT & ES & \numprint{0.000} & \numprint{0.080} \\
\cdashline{2-6}\noalign{\vskip \belowrulesep}
& OT & AT & EM & \numprint{0.000} & \numprint{-0.011} \\
& OT & AT & ES & \numprint{0.000} & \numprint{-0.011} \\
\midrule
\multirow{4.4}{*}{CodeGPT}
& NT & OT & EM & \numprint{0.000} & \numprint{0.021} \\
& NT & OT & ES & \numprint{0.000} & \numprint{0.026} \\
\cdashline{2-6}\noalign{\vskip \belowrulesep}
& OT & AT & EM & \numprint{0.000} & \numprint{-0.010} \\
& OT & AT & ES & \numprint{0.000} & \numprint{-0.011} \\
\midrule
\multirow{3.4}{*}{InCoder}
& NT & OT & EM & \numprint{0.000} & \numprint{0.012} \\
& NT & OT & ES & \numprint{0.000} & \numprint{0.021} \\
\cdashline{2-6}\noalign{\vskip \belowrulesep}
& OT & AT & ES & \numprint{0.000} & \numprint{0.003} \\
\bottomrule
\end{tabular}
\end{table}

\subsubsection{Token Completion}

\begin{table}[tb]
\caption{RQ1: Impact of Type Annotations (Token Completion, Data without Comments)}
\label{table:rq1_metrics_token}
\centering
\begin{tabular}{llrr}
\toprule
Model & Types & EM & ES\\
\midrule
\multirow{3}{*}{UniXcoder}
& NT & \textbf{\numprint{78.28}} & \textbf{\numprint{81.36}}\\
& OT & \numprint{70.80} & \numprint{73.88}\\
& AT & \numprint{71.81} & \numprint{74.74}\\

\midrule
\multirow{3}{*}{CodeGPT}
& NT & \textbf{\numprint{78.47}} & \textbf{\numprint{82.02}}\\
& OT & \numprint{75.35} & \numprint{78.98}\\
& AT & \numprint{76.42} & \numprint{79.93}\\
\midrule
\multirow{3}{*}{InCoder}
& NT & \textbf{\numprint{51.46}} & \textbf{\numprint{57.05}}\\
& OT & \numprint{50.94} & \numprint{56.61}\\
& AT & \numprint{51.15} & \numprint{56.64}\\
\bottomrule
\end{tabular}
\end{table}

Table~\ref{table:rq1_metrics_token} presents
how each model performed on the token completion task
on comment-less code.
We only report Exact Match and Edit Similarity, as all other metrics are sequence-level metrics that do not work on single tokens.
We perform the Wilcoxon Signed Rank test on these results, as displayed in Table~\ref{table:results_rq1_p_values_token}.

\textbf{Removing type annotations} once again leads to the best performance.
The performance gain is statistically significant for all models, with effect sizes similar to those found for line-level completion.

\textbf{Adding type annotations} improves auto-completion performance.
This outperformance is statistically significant for all metrics for UniXcoder and CodeGPT, and statistically significant only for Exact Match for InCoder.
The effect sizes are small across all models.

Overall, the results for token completion are similar to the results for line completion. Removing type annotations leads to a larger performance gain than adding type annotations. The reason for this is not clear from our experiments alone, however, we theorize that this could indicate that the tested models are better at interpreting source code than type annotations, making it worthwhile to remove type annotations to make more space for source code tokens.

\begin{table}[tb]
\caption{RQ1: Impact of Type Annotations (Token Completion, data without comments, 
Wilcoxon Signed Rank $p$-values, Cliff's Delta)}
\label{table:results_rq1_p_values_token}
\centering
\begin{tabular}{llllrr}
\toprule
Model & Types 1 & Types 2 & Metric & $p$ & $\delta$ \\
\midrule
\multirow{4.4}{*}{UniXcoder}
& NT & OT & EM & \numprint{0.000} & \numprint{0.075} \\
& NT & OT & ES & \numprint{0.000} & \numprint{0.081} \\
\cdashline{2-6}\noalign{\vskip \belowrulesep}
& OT & AT & EM & \numprint{0.000} & \numprint{-0.010} \\
& OT & AT & ES & \numprint{0.000} & \numprint{-0.010} \\
\midrule
\multirow{4.4}{*}{CodeGPT}
& NT & OT & EM & \numprint{0.000} & \numprint{0.025} \\
& NT & OT & ES & \numprint{0.000} & \numprint{0.027} \\
\cdashline{2-6}\noalign{\vskip \belowrulesep}
& OT & AT & EM & \numprint{0.000} & \numprint{-0.011} \\
& OT & AT & ES & \numprint{0.000} & \numprint{-0.011} \\
\midrule
\multirow{3.4}{*}{InCoder}
& NT & OT & EM & \numprint{0.000} & \numprint{0.004} \\
& NT & OT & ES & \numprint{0.000} & \numprint{0.004} \\
\cdashline{2-6}\noalign{\vskip \belowrulesep}
& OT & AT & EM & \numprint{0.000} & \numprint{-0.002} \\
\bottomrule
\end{tabular}
\end{table}

    \textbf{Answer to RQ1}:
    All models perform best on untyped code across nearly all metrics.
    The Wilcoxon Signed Rank test shows the performance difference 
    between untyped (TS704-NT) and type-annotated code (TS704-OT) is significant.
    This suggests that type annotations 
    do not necessarily enhance auto-completion models' ability to interpret and complete code.
    Additionally, performance on code with a high type explicitness (TS704-AT) 
    is significantly better than performance on code with a normal type explicitness ratio (TS704-OT).
    This could suggest that the irregular nature of type annotations 
    in gradually-typed languages may make it more difficult 
    for language models to interpret optionally-typed languages.
    Comparing effect sizes indicates that it is a better option
    to strip (rather than add) type annotations from TypeScript code
    to improve auto-completion performance.
    We theorize that these observations suggest that source code provides more value to these models, rather than type annotations. 
    Consequently, removing type annotations to allow more source code to be used by the models can lead to increased performance.

\subsection{RQ2: Impact of Comments}

We run UniXcoder, CodeGPT, and InCoder against the test sets
of TS704-NT with different types of comments 
to determine whether specific types of comments influence auto-completion performance.
Similarly, we use the Wilcoxon Signed Rank test 
to determine significance, and consider both line and token completion.

\subsubsection{Line Completion}
Table~\ref{table:results_rq2} shows how each model performed on
code containing different types of comments.
In this table, and the tables thereafter, 
the \textit{CMT} column indicates which types of comments are present in the code.
Possible values are 
\textbf{NC} (No Comments), 
\textbf{SL} (Single-Line comments), 
\textbf{ML} (Multi-Line comments), 
\textbf{DB} (Doc-Block comments), and 
\textbf{AC} (All Comments).

We conduct the Wilcoxon Signed Rank test to determine whether
any observations are significant.
We use performance on comment-less data as a baseline, 
and compare it to performance on all other types of comments.
We also compute Cliff's delta.
The results are shown in Table~\ref{table:results_rq2_p_values}.
The B4, RL, and MR metrics are once more omitted for brevity,
as these were significant in nearly all cases, 
and show great correlation to EM and ES.

Preserving \textbf{All Comments or Multi-Line Comments} leads to the best performance for all models and all metrics.
This implies that the information embedded in multi-line comments leads to the best understanding of the source code.
UniXcoder, CodeGPT, and InCoder are all (pre-)trained on corpora containing English text, 
which could explain their outperformance on code with multi-line comments relative 
to their performance on code without such comments.
The outperformance on these types of comments is statistically significant for all three models, albeit with small effect sizes.
Keeping All Comments versus only multi-line comments have similar effects for UniXcoder and CodeGPT, but for InCoder specifically, multi-line comments have a larger effect size.

Preserving solely \textbf{Doc-Block Comments} does not lead to substantial nor significant performance gains, despite Doc-Blocks being a type of multi-Line comment. This reinforces that the natural language descriptions inside multi-line comments cause performance enhancements, rather than Doc-Block information such as argument types and purposes.

Keeping only \textbf{Single-Line Comments} generally does not cause performance enhancements like preserving only multi-line comments.
This could be explained by the differences in comment length: single-line comments are generally much smaller than multi-line comments.
For UniXcoder, Single-Line comments appear to cause a very small, yet statistically significant performance increase.
For InCoder, however, these comments have a statistically significant negative effect, one more with a small effect size.
Single-Line comments do not significantly impact the performance of CodeGPT.

Overall, different types of comments have a statistically significant effect on line-level auto-completion performance, albeit with relatively small effect sizes.
multi-line comments appear to have the largest positive effect, indicating that the three models are most capable at interpreting these types of comments.

\begin{table}[tb]
\caption{RQ2: Impact of Comments on Line Completion, TS704-NT}
\label{table:results_rq2}
\centering
\begin{tabular}{llrrrrr}
\toprule
Model & CMT & EM & ES & B4 & RL & MR \\
\midrule
\multirow{5}{*}{UniXcoder}
& NC & \numprint{65.32} & \numprint{79.62} & \numprint{61.63} & \numprint{81.32} & \numprint{65.91}\\
& SL & \numprint{65.84} & \numprint{80.30} & \numprint{62.16} & \numprint{82.03} & \numprint{66.41}\\
& ML & \textbf{\numprint{69.28}} & \numprint{82.99} & \numprint{63.78} & \numprint{84.35} & \numprint{67.65}\\
& DB & \numprint{65.17} & \numprint{79.56} & \numprint{61.55} & \numprint{81.24} & \numprint{65.78}\\
& AC & \numprint{69.27} & \textbf{\numprint{83.09}} & \textbf{\numprint{63.88}} & \textbf{\numprint{84.46}} & \textbf{\numprint{67.76}}\\
\midrule
\multirow{5}{*}{CodeGPT}
& NC & \numprint{63.39} & \numprint{80.38} & \numprint{62.08} & \numprint{82.51} & \textbf{\numprint{67.03}}\\
& SL & \numprint{62.85} & \numprint{79.97} & \numprint{61.72} & \numprint{82.08} & \numprint{66.68}\\
& ML & \textbf{\numprint{66.90}} & \textbf{\numprint{82.56}} & \textbf{\numprint{62.82}} & \textbf{\numprint{84.26}} & \numprint{66.95}\\
& DB & \numprint{63.65} & \numprint{80.42} & \numprint{62.10} & \numprint{82.48} & \numprint{67.02}\\
& AC & \numprint{66.45} & \numprint{82.29} & \numprint{62.64} & \numprint{84.01} & \numprint{66.78}\\
\midrule
\multirow{5}{*}{InCoder}
& NC & \numprint{33.26} & \numprint{56.56} & \numprint{43.51} & \numprint{59.71} & \numprint{51.09}\\
& SL & \numprint{32.30} & \numprint{55.86} & \numprint{43.09} & \numprint{59.05} & \numprint{50.68}\\
& ML & \textbf{\numprint{34.89}} & \textbf{\numprint{57.78}} & \textbf{\numprint{44.13}} & \textbf{\numprint{60.67}} & \textbf{\numprint{51.37}}\\
& DB & \numprint{33.27} & \numprint{56.49} & \numprint{43.49} & \numprint{59.56} & \numprint{50.99}\\
& AC & \numprint{33.99} & \numprint{57.02} & \numprint{43.64} & \numprint{59.91} & \numprint{50.88}\\
\bottomrule
\end{tabular}
\end{table}

\begin{table}[tb]
\caption{RQ2: Impact of Comments on Line Completion, TS704-NT (Wilcoxon Signed Rank $p$-values, Cliff's Delta)}
\label{table:results_rq2_p_values}
\centering
\begin{tabular}{llllrr}
\toprule
Model & CMT 1 & CMT 2 & Metric & $p$ & $\delta$ \\
\midrule
\multirow{9.2}{*}{UniXcoder}
& NC & SL & EM & \numprint{0.000} & \numprint{-0.003} \\
& NC & SL & ES & \numprint{0.000} & \numprint{-0.005} \\
\cdashline{2-6}\noalign{\vskip \belowrulesep}
& NC & ML & EM & \numprint{0.000} & \numprint{-0.031} \\
& NC & ML & ES & \numprint{0.000} & \numprint{-0.036} \\
\cdashline{2-6}\noalign{\vskip \belowrulesep}
& NC & DB & EM & \numprint{0.000} & \numprint{0.003} \\
& NC & DB & ES & \numprint{0.003} & \numprint{0.003} \\
\cdashline{2-6}\noalign{\vskip \belowrulesep}
& NC & AC & EM & \numprint{0.000} & \numprint{-0.032} \\
& NC & AC & ES & \numprint{0.000} & \numprint{-0.039} \\
\midrule
\multirow{5.8}{*}{CodeGPT}
& NC & ML & EM & \numprint{0.000} & \numprint{-0.020} \\
& NC & ML & ES & \numprint{0.000} & \numprint{-0.022} \\
\cdashline{2-6}\noalign{\vskip \belowrulesep}
& NC & DB & EM & \numprint{0.019} & \numprint{-0.001} \\
\cdashline{2-6}\noalign{\vskip \belowrulesep}
& NC & AC & EM & \numprint{0.000} & \numprint{-0.020} \\
& NC & AC & ES & \numprint{0.000} & \numprint{-0.023} \\
\midrule
\multirow{6.8}{*}{InCoder}
& NC & SL & EM & \numprint{0.000} & \numprint{0.008} \\
& NC & SL & ES & \numprint{0.000} & \numprint{0.009} \\
\cdashline{2-6}\noalign{\vskip \belowrulesep}
& NC & ML & EM & \numprint{0.000} & \numprint{-0.017} \\
& NC & ML & ES & \numprint{0.000} & \numprint{-0.019} \\
\cdashline{2-6}\noalign{\vskip \belowrulesep}
& NC & AC & EM & \numprint{0.000} & \numprint{-0.010} \\
& NC & AC & ES & \numprint{0.000} & \numprint{-0.010} \\
\bottomrule
\end{tabular}
\end{table}

\subsubsection{Token Completion}

\begin{table}[tb]
\caption{RQ2: Impact of Comments on Token Completion, TS704-NT}
\label{table:results_rq2_token}
\centering
\begin{tabular}{llrr}
\toprule
Model & CMT & EM & ES \\
\midrule
\multirow{5}{*}{UniXcoder}
& NC & \numprint{78.28} & \numprint{81.36}\\
& SL & \numprint{78.83} & \numprint{81.91}\\
& ML & \textbf{\numprint{82.35}} & \textbf{\numprint{85.23}}\\
& DB & \numprint{78.35} & \numprint{81.39}\\
& AC & \numprint{82.30} & \textbf{\numprint{85.23}}\\
\midrule
\multirow{5}{*}{CodeGPT}
& NC & \numprint{78.47} & \numprint{82.02}\\
& SL & \numprint{77.85} & \numprint{81.44}\\
& ML & \textbf{\numprint{80.99}} & \textbf{\numprint{84.31}}\\
& DB & \numprint{78.42} & \numprint{81.86}\\
& AC & \numprint{80.66} & \numprint{84.05}\\
\midrule
\multirow{5}{*}{InCoder}
& NC & \numprint{51.46} & \numprint{57.05}\\
& SL & \numprint{50.45} & \numprint{56.21}\\
& ML & \textbf{\numprint{52.47}} & \textbf{\numprint{58.06}}\\
& DB & \numprint{51.32} & \numprint{56.92}\\
& AC & \numprint{51.47} & \numprint{57.22}\\
\bottomrule
\end{tabular}
\end{table}

\begin{table}[tb]
\caption{RQ2: Impact of Comments on Token Completion, TS704-NT 
(Wilcoxon Signed Rank $p$-values, Cliff's Delta)}
\label{table:results_rq2_p_values_token}
\centering
\begin{tabular}{llllrr}
\toprule
Model & CMT 1 & CMT 2 & Metric & $p$ & $\delta$ \\
\midrule
\multirow{6.8}{*}{UniXcoder}
& NC & SL & EM & \numprint{0.000} & \numprint{-0.004} \\
& NC & SL & ES & \numprint{0.000} & \numprint{-0.004} \\
\cdashline{2-6}\noalign{\vskip \belowrulesep}
& NC & ML & EM & \numprint{0.000} & \numprint{-0.032} \\
& NC & ML & ES & \numprint{0.000} & \numprint{-0.034} \\
\cdashline{2-6}\noalign{\vskip \belowrulesep}
& NC & AC & EM & \numprint{0.000} & \numprint{-0.033} \\
& NC & AC & ES & \numprint{0.000} & \numprint{-0.036} \\
\midrule
\multirow{6.8}{*}{CodeGPT}
& NC & SL & EM & \numprint{0.006} & \numprint{0.002} \\
& NC & SL & ES & \numprint{0.003} & \numprint{0.002} \\
\cdashline{2-6}\noalign{\vskip \belowrulesep}
& NC & ML & EM & \numprint{0.000} & \numprint{-0.017} \\
& NC & ML & ES & \numprint{0.000} & \numprint{-0.019} \\
\cdashline{2-6}\noalign{\vskip \belowrulesep}
& NC & AC & EM & \numprint{0.000} & \numprint{-0.018} \\
& NC & AC & ES & \numprint{0.000} & \numprint{-0.019} \\
\midrule
\multirow{6.8}{*}{InCoder}
& NC & SL & EM & \numprint{0.000} & \numprint{0.009} \\
& NC & SL & ES & \numprint{0.000} & \numprint{0.009} \\
\cdashline{2-6}\noalign{\vskip \belowrulesep}
& NC & ML & EM & \numprint{0.000} & \numprint{-0.011} \\
& NC & ML & ES & \numprint{0.000} & \numprint{-0.012} \\
\cdashline{2-6}\noalign{\vskip \belowrulesep}
& NC & AC & EM & \numprint{0.002} & \numprint{-0.002} \\
& NC & AC & ES & \numprint{0.000} & \numprint{-0.004} \\
\bottomrule
\end{tabular}
\end{table}

Table~\ref{table:results_rq2_token} 
shows the token completion performance
of all models on code with different types of comments.
We only report Exact Match and Edit Similarity, as the other metrics are not applicable to single tokens.
We once more perform the Wilcoxon Signed Rank Test on the results, as shown in Table~\ref{table:results_rq2_p_values_token}.
Overall, the results are similar to the results observed for line completion.
Preserving \textbf{All Comments or Multi-Line Comments} once more leads to the best performance.
This outperformance is significant and has the largest effect size for all three models.
Once more, UniXcoder and CodeGPT have similar effect sizes when preserving multi-line and all comments,
whereas InCoder appears to benefit more when preserving only multi-line comments.

Preserving only \textbf{Doc-Block Comments} once again does not lead to significant performance improvements, despite doc-block comments being a type of multi-line comment.
Keeping only \textbf{Single-Line} comments significantly affects performance for all models.
For UniXcoder and InCoder there are slightly positive and negative effect sizes respectively, which was also the case for line completion.
For CodeGPT, we observe a slight, statistically significant, negative effect, which was not observed during line completion.

Overall, the results for token completion are consistent with the results for line completion: different types of comments influence auto-completion performance in different ways, with small effect sizes. Multi-Line comments appear to have the largest positive effect for the three models.

    \textbf{Answer to RQ2}:
    All three models perform best on a code containing either all comments or solely multi-line comments.
    The presence of doc-block comments (which is a type of multi-line comment) 
    does not cause a significant performance increase, 
    suggesting that the value of multi-line comments comes 
    from the natural language embedded in them.
    Additionally, single-line comments do not always appear 
    to have the same effect: 
    UniXcoder slightly benefits from the presence of single-line comments, 
    but CodeGPT and InCoder experience performance degradation when present.
    While code containing multi-line comments performs best,
    the effect size is relatively small.
    Nevertheless, these types of comments 
    do provide some value to code completion models.
    The results suggest that the three code completion models can
    adequately interpret natural language descriptions contained
    in multi-line comments.
    Other comment types do not appear to further enhance auto-completion performance, suggesting that these comments can be omitted from the input without sacrificing performance.

\subsection{Discussion and Recommendations}
The experiments' results indicate that 
the observed differences in auto-completion performance 
are statistically significant, 
hence they are not random. 
However, the effect sizes are small, 
which can indicate limited impact in practice.

In this work, we focused on TypeScript. 
More studies are required to generalize our findings 
including investigating whether the observed results 
apply to other programming languages or not.
Python3 is a suitable candidate, 
as it is an optionally-typed language.
However, it lacks a type inference engine
that is present in the TypeScript compiler, 
which makes this more challenging.
Additionally, experiments with different splitting policies 
(e.g., repository-based instead of file-based), 
other datasets, and more strictly de-duplication policies 
could reinforce our findings.

The results hint at the removal of type annotations being beneficial,
which may suggest that current LLMs do not need additional clues 
such as type annotations to aid code understanding.
This is specifically important as the input size of LLMs is generally limited -- removing type annotations can make space for more important information.
Examples of alternative types of contextual information that could be of more use to these models are 
1) available function signatures, 
2) local folder structure, 
3) previous (correct) predictions, and 
4) simply more code context.
Our results indicate that predominantly multi-line comments are of importance for code completions.
Removing all other types of comments can similarly create extra input space for other types of useful information.

In general, we believe that helping recent LLMs understand the input syntactically may not be the best way of advancing state-of-the-art models with millions or billions of parameters.
Instead, providing different types of contextual information may help widen the scope of code completion models.
Future research is needed to confirm how different types of contextual clues may be valuable to LLMs used for code-related tasks.

\subsection{Threats to the Validity}
\label{sec:threats}
We categorize threats to the validity of our study
into three groups, namely
internal, external, and construct threats.

\textbf{Threats to internal validity}: 
relate to the parameters affecting the performance of the model,
factors unintentionally influencing the results,
and errors in the implementations.
We intentionally changed
type annotations and comments 
to test their relationship with auto-completion performance.
Variables involved in this work are 
fine-tuning hyper-parameters and metric-related parameters.
Hyper-parameter tuning was performed before fine-tuning UniXcoder
and CodeGPT, after which the hyper-parameters that lead to
the best accuracy were kept constant throughout the experiments.
This ensures that no observed differences between fine-tuned models
can be attributed to a difference in the fine-tuning configuration.
The same applies to parameters relating to metrics; 
the tokenization of input sequences required for BLEU, \mbox{ROUGE-L} and METEOR 
was always done the same way, according to the ECMAScript lexical grammar specification~\cite{ESTokenSpec}.
Moreover, our analysis relies on 
pre-existing code published by~\citeauthor*{guo2022unixcoder} 
and~\citeauthor*{Lu2021}
to make the evaluation of UniXcoder and CodeGPT reproducible~\cite{guo2022unixcoder, Lu2021}.
This perfectly demonstrates the importance of reproducibility.
We assure reproducibility by publishing all resources 
required to conduct the experiment, 
including the source code and datasets.
Additionally, we provide fine-tuned models
for UniXcoder and CodeGPT for all three
type explicitness settings, trained on data
containing all comment types.

\textbf{Threats to external validity}:
relate to factors that could 
affect the generalizability of our findings.
In this study, 
the datasets used greatly impact the generalizability of results.
We create our own dataset, 
\textit{TS704-OT},
based on which we create two more datasets, 
\textit{TS704-AT} and \textit{TS704-NT},
through adding and removing type annotations.
To prevent skewed results, 
we removed all repositories in our dataset that 
were also used to train UniXcoder, CodeGPT, or InCoder.
However, our datasets themselves were not de-duplicated, 
which could lead to overly-optimistic results 
caused by the overlap of the train and test sets~\cite{allamanis2019adverse}.
Our duplication measurements indicate 
that there is only 1\% exact duplication within our dataset, 
and 7\% near-duplication. 
We believe this duplication is relatively small 
and aligns with real scenarios 
as in practice developers tend to reuse code frequently.
Additionally, the datasets were split by file,
as opposed to by project, or by repository.
This could cause artificially high measurements 
as projects generally share patterns or variable names~\cite{leclair2019neural}.
Hence, different splitting policies should be explored in the future.
Future work could also incorporate different optionally-typed languages, 
and different datasets to further support our findings.

\textbf{Threats to construct validity}: 
relate to the validity of the measurements performed.
We use several commonly-used metrics in the NLP field~\cite{Vaswani2017, Lu2021, guo2022unixcoder, Wang2021}.
Together, these metrics give a broad perspective 
on the performance of models,
as each metric measures performance in a different way.
BLEU, ROUGE-L, and METEOR 
all require parameters or tokenization.
Slight differences in the way these metrics are applied 
can wildly change results 
and differences in parameters 
make the metrics incomparable~\cite{post2018call}.
This highlights the importance of 
indicating exactly how each metric is used.
The specification of datasets is similarly important~\cite{post2018call} 
and plays a big part in reproducibility.
For these reasons the processes applied 
to obtain the results are described in detail and transparently, 
such that it is clear how the data should be interpreted, 
and whether it is comparable to data provided in other studies.
To ensure that all data reported 
are comparable, 
we used the same distribution of
files in the train, validation, and test sets
for all experiments.

\section{Conclusion}\label{sec:conclusion}

Source code tokens are not the only sources 
of information for exploiting the context 
when using code LMs.
Optional type annotations 
and natural language text in the form of comments 
can add valuable information to the source code. 
However, not all this extra information may be
useful for improving language understanding
capabilities of auto-completion models.
In this work, we investigated the impact of
these sources of additional information 
when leveraged by three recent LMs for source code.
Our results show that
not all optional information channels are valuable
to the three models.
Namely, type annotations are shown to negatively affect these models,
whilst their performance is enhanced by the presence of multi-line comments.
These observations are statistically significant for all models
and nearly all metrics (albeit with small effect sizes).
Hence, the community should take these findings into account 
when selecting the best model for their tasks.
For instance, exchanging type annotations and non-multi-line comments 
in the input of these models
for other more beneficial types of contextual information 
can be helpful to these models.
Future research can investigate how different types of contextual clues 
can impact the performance of LLMs for code.

\section{Data Availability Statement}
Our source code, dataset, and select fine-tuned models are publicly available.\footnote{\url{https://github.com/AISE-TUDelft/ContextualDataCodeCompletion}; \url{https://huggingface.co/AISE-TUDelft/CodeGPT-TS-Multi-Untyped}; \url{https://huggingface.co/AISE-TUDelft/UniXcoder-TS-Multi-Untyped}}

\printbibliography

\end{document}